%% file: main.tex
\documentclass[11pt]{article}

\usepackage[preprint]{acl}

\usepackage{tikz}
\usepackage{xcolor}
\usepackage{times}
\usepackage{latexsym}

\usetikzlibrary{shapes.geometric, arrows.meta, positioning, fit, calc, backgrounds}

\newcommand{\appref}[1]{\hyperref[#1]{Appendix~\ref*{#1}}}

\input{preamble}

\title{Agent Skill Evaluation and Evolution: Frameworks and Benchmarks}
\author{Author Name \\
  Affiliation \\
  \texttt{email@institution.org}}

\author{
 \textbf{Kexin Ding\textsuperscript{1,2,3}},
 \textbf{Yang Zhou\textsuperscript{1,2}},
 \textbf{Can Jin\textsuperscript{1,2}},
 \textbf{Feng Tong\textsuperscript{4}},
 \\
 \textbf{Mu Zhou\textsuperscript{1}},
 \textbf{Dimitris N. Metaxas\textsuperscript{1,2,3}}\
\\
\\
 \textsuperscript{1}Rutgers, the State University of New Jersey,
\\
 \textsuperscript{2}Center for Computational Biomedicine, Imaging and Modeling (CBIM),
\\
\textsuperscript{3}Rutgers Artificial Intelligence and Data Science (RAD) Collaboratory,
\\
 \textsuperscript{4}University of North Carolina at Charlotte
\\
 \small{
   \textbf{Correspondence:} \href{mailto:email@domain}{dnm@cs.rutgers.edu}
 }
}

\begin{document}
\maketitle

\begin{abstract}
The growth of \emph{agent skills} has transformed how agentic systems are built, evaluated, and deployed. As skill libraries continue to scale, rigorous evaluation becomes critical to ensuring their utility, quality, and safety in real-world applications. Consequently, the field is undergoing an emerging paradigm shift from isolated skill creation to automated, evaluation-driven skill evolution. In this survey, we systematically examine the landscape of skill evolution and evaluation beyond foundational skill creation. We categorize evolution into four distinct paradigms, spanning execution feedback, trajectory distillation, compression, and reinforcement learning, showing how each element contributes to improving skill utility and reliability. We also provide an analysis of six skill-centric benchmark categories, identifying structural gaps in benchmark coverage, trade-offs, and metric richness to advance skill research. Finally, we identify open directions for building skill ecosystems that are generalizable, efficient, and verifiably safe. The project URL is \url{https://github.com/Cassie07/AgentSkill_Survey}

%For each theme, we define the core problem and motivation, synthesize key findings across papers, and highlight contrasts and complementarities.
\end{abstract}

\input{sections/01_introduction}

\input{sections/02_skill_collection}
\input{sections/04_skill_evolution}
\input{sections/05_benchmarks}
\input{sections/06_discussion}

\input{sections/07_conclusion}

% \clearpage
% \listoftables

% \bibliography{references}
\bibliography{ref}

\appendix
\input{sections/09_supp_skill_usage}
\input{sections/09_supp_general_domain}

\input{sections/10_analysis_comparison}
\end{document}

%% file: preamble.tex
\usepackage[T1]{fontenc}
\usepackage[utf8]{inputenc}
\usepackage{microtype}
\usepackage{graphicx}
\usepackage{booktabs}
\usepackage{longtable}
\usepackage{tabularx}
\usepackage{dblfloatfix}
\usepackage[section]{placeins}
\usepackage{amsmath,amssymb}
\usepackage[nameinlink,capitalize,noabbrev]{cleveref}

\usepackage[table]{xcolor}
\usepackage{tabularx}
\usepackage{booktabs}

\usepackage{longtable}
\usepackage{array}
\usepackage{colortbl}
\usepackage[table]{xcolor} % if not already loaded
\usepackage{makecell} % for \thead

\usepackage{fontawesome5}
\usepackage{xurl}

\usepackage{tikz}
\usetikzlibrary{positioning,arrows.meta,fit,backgrounds,calc,arrows.meta, positioning, shapes.geometric}

\definecolor{catFeedback}{RGB}{255,229,204} % light orange
\definecolor{catDistill}{RGB}{217,234,211}  % light green
\definecolor{catCompAug}{RGB}{207,226,243}  % light blue
\definecolor{catRL}{RGB}{230,217,243}       % light purple

\definecolor{catUtility}{RGB}{230,240,255}   % light blue
\definecolor{catGen}{RGB}{255,250,205}       % light yellow
\definecolor{catRetrieval}{RGB}{250,235,215} % light orange
\definecolor{catSafety}{RGB}{240,225,250}    % light purple
\definecolor{catSWE}{RGB}{230,245,225}       % light green
\definecolor{catWild}{RGB}{255,230,230}      % light red

\DeclareRobustCommand{\catbox}[1]{%
  {\setlength{\fboxsep}{0pt}\setlength{\fboxrule}{0.2pt}%
   \fcolorbox{black!30}{#1}{\rule{0pt}{1.4ex}\rule{1.4ex}{0pt}}}}

\DeclareRobustCommand{\catTag}[3]{%
  \tikz[baseline=(t.base)]{%
    \node[fill=#1, rounded corners=2.5pt,
          inner xsep=4pt, inner ysep=0.6pt,
          text depth=0.35ex,
          draw=#1!65!black, line width=0.2pt,
          font=\small\bfseries]
    (t) {\strut #2\,#3};%
  }%
}

\DeclareRobustCommand{\catFeedbackTag}{\catTag{catFeedback}{\faSync}{Execution Feedback}}
\DeclareRobustCommand{\catDistillTag}{\catTag{catDistill}{\faRoute}{Trajectory Distillation}}
\DeclareRobustCommand{\catCompAugTag}{\catTag{catCompAug}{\faCompress}{Compression and Augmentation}}
\DeclareRobustCommand{\catRLTag}{\catTag{catRL}{\faTrophy}{Reinforcement Learning}}

\DeclareRobustCommand{\catUtilityTag}{\catTag{catUtility}{\faWrench}{Skill Utility Benchmarks}}
\DeclareRobustCommand{\catGenTag}{\catTag{catGen}{\faSeedling}{Skill Generation Benchmarks}}
\DeclareRobustCommand{\catRetrievalTag}{\catTag{catRetrieval}{\faSearch}{Skill Retrieval \& Routing Benchmarks}}
\DeclareRobustCommand{\catSafetyTag}{\catTag{catSafety}{\faLock}{Skill Safety Auditing Benchmarks}}
\DeclareRobustCommand{\catSWETag}{\catTag{catSWE}{\faLaptopCode}{Software Engineering Benchmark}}
\DeclareRobustCommand{\catWildTag}{\catTag{catWild}{\faGlobe}{Real-World Environment Benchmarks}}

%% file: sections/01_introduction.tex
\section{Introduction}
\label{sec:intro}

% TODO: Industry challenges

% Figure: SKILL and agent collaboration (Overview)

% PARAGRAPH 1: Three Paradigms + Challenges of Each 

% Large language models (LLMs) agents exhibit remarkable capabilities to perceive environments, plan multi-step actions, and interact with external tools~\cite{xi2025rise}. %Such interaction capabilities of LLM agents could be attributed to their promising abilities in natural language understanding and generation, as well as in complex multi-step reasoning and code synthesis~\cite{brown2020language,achiam2023gpt,zhao2023survey}. %Despite the rapid progress of LLM agents, building reliable agent systems that accumulate and reuse domain knowledge across tasks remains an open challenge. 

% (Comment: can be moved to the introduction) Skills empower LLM agents in handling complex tasks, covering areas such as coding, data analysis, document writing, computer use, and robot control~\cite {jiang2026sok,li2026organizing,zheng2026skillrouter}. 

Agent skills equip LLM agents with domain-specific knowledge at inference time, enabling agents to perceive and interact with environments through diverse external tools~\cite{zhang2025equipping}. Unlike prompt engineering~\cite{wei2022chain,brown2020language}, agent skills encode reusable, portable, multi-step solutions that guide agents to address complex tasks through coordinated decision sequences, thereby substantially reducing tedious manual effort.

%In practice, the skill is implemented as a \texttt{SKILL.md} file~\cite{zhang2025equipping}, bundling instructions, workflow guidance, executable scripts, tool interaction patterns, and metadata into a standardized package ~\cite{zhang2025equipping,xu2026agent}. 
% ~\cite{wei2022chain,kojima2022large}.  

% Meanwhile, skills supporting version control and sharing, allowing them to be loaded by any compatible agent harness, such as Claude Code, Codex, and Gemini CLI.

As the scale and diversity of agent skills continue to grow~\cite{liang2026skillnet,li2026skillsbench}, the absence of robust evaluation frameworks has become a critical bottleneck for skill-guided agent deployment. Meanwhile, diverse skills make the manual refinement inherently infeasible, a challenge further compounded by the lack of evolution approaches to capture the real-world feedback~\cite{zhang2026coevoskills}. An outdated or unsafe skill can propagate errors across downstream tasks, turning skill assessment into an open problem of diagnosis, maintenance, and alignment. It is thus essential to establish automated and continuous mechanisms for agent skills, rather than relying on static pipelines, to ensure skills are generalizable across tasks and verifiably safe for public use. In this survey, we position skill evolution and evaluation as the central focus of this emerging paradigm (Figure~\ref{fig1}). Our survey is intentionally not an exhaustive census but a focused review of skill evolution and evaluation on a curated subset of skill literature. We searched via Google Scholar using "agent skill", "skill", and "skill evolution" as keywords, covering works through May 2026, and the inclusion criterion centered on our two main themes. Concretely, we introduce a four-paradigm taxonomy of skill evolution strategies (\autoref{sec:2.3}). We gain insights into designing evolution strategies towards enhancing skill creation, utility, and refinement with fewer human efforts. We further provide a critical analysis of skill-centric benchmarks (\autoref{sec:2.5}) to assess their opportunities in multimodal skills, trajectory distillation, and skill security towards better real-world agent deployment.

%% file: sections/02_skill_collection.tex
\section{What is Skill?}
\label{sec:2.1}

\begin{figure*}[t!]
\centering
\includegraphics[width=\textwidth]{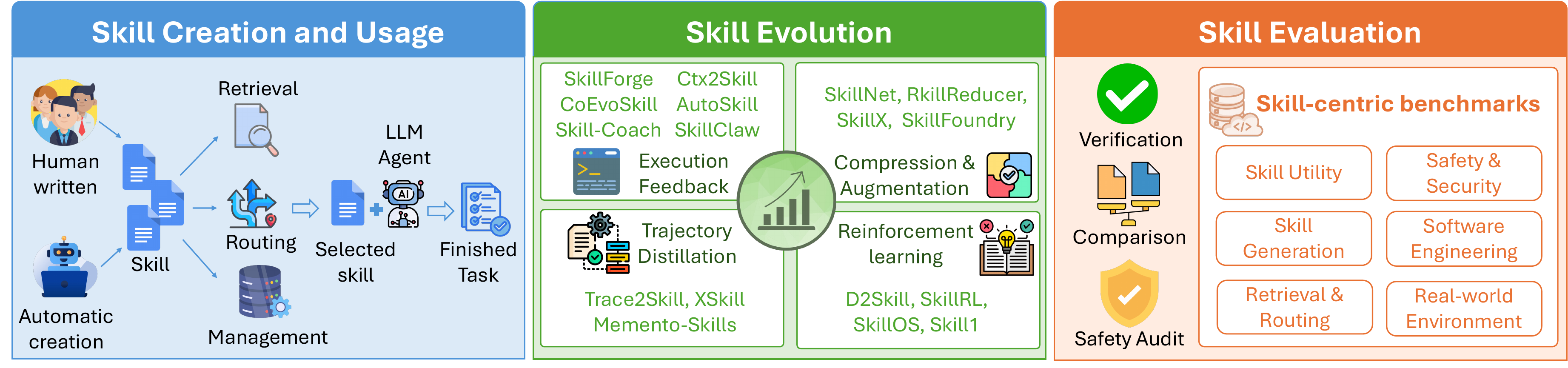}
\caption{We map the landscape of agent skill evolution strategies (\S3) through comparative analysis and design recommendations. We offer evaluation insights (\S4) through structural gaps and benchmark limitations, and open challenges (\S5) for robust real-world skill deployment.}
\label{fig1}
\end{figure*}
% ~\cite{jiang2026sok}

An agent skill is a structured package~\cite{li2026skillsbench}:
$\mathcal{S} = (C,\,\pi,\,T,\,\mathcal{R})$, where $C: \mathcal{O} \times \mathcal{G} \rightarrow \{0,1\}$ is the condition, mapping the agent observation ($\mathcal{O}$) and goal ($\mathcal{G}$) to the skill relevance; $\pi$ is the execution policy to encode the procedures; $T$ is the termination criterion, specifying when skill execution is completed; and $\mathcal{R}$ is the reusable interface to indicate the composition with other skills.
% , and distributed across agent harnesses. 

Human-authored skills encapsulate domain expertise as machine-interpretable procedural knowledge~\cite{li2026skillsbench}. To expedite this process, automated skill creation enables an agent to generate skills with less human-written effort. For example, Skill Creator~\cite{anthropic_skills} could automatically create a full skill directory and test cases with minimal human text description. Similarly, Voyager~\cite{wang2023voyager} creates skills as the executable code, including proposing tasks, refining code via environment feedback, self-verification, and updating the skill library. To better create reusable skills, reinforcement learning (RL) is integrated into the training loop, where rewards earned from reusing a skill on later tasks are propagated back to update the policy. Inspired by Group Relative Policy Optimization (GRPO)~\cite{shao2024deepseekmath}, SAGE~\cite{wang2025reinforcement} leverages the reusable rewards from group tasks, encouraging the agent to learn and create reusable skills. ARISE~\cite{li2026arise} preserves successful reasoning patterns to train agents toward generating reusable skills, overcoming GRPO's limitation of treating rollouts independently. 

Efficient skill usage strategies involve retrieval, routing, and management. For each task, it is common that the agent cannot load all potential skills to assess their usability due to excessive time and token use. To address, (a) Retrieval determines a small set of skills from a large skill pool; (b) Routing efficiently decides which skill should be executed at which step after retrieval; (c) Management keeps the skills organized, up-to-date, and safe to use (\appref{supp_skill_usage}). These usage mechanisms establish the foundation upon which evolution and evaluation frameworks operate as outlined below.

%% file: sections/04_skill_evolution.tex
\section{Skill Evolution}
\label{sec:2.3}
%add some formal technical definitions in proper places
%  To enable skills aligned with unseen tasks, it is increasingly important to explore strategies to improve existing skills at scale. 

Skill evolution is a continuous process to improve skill quality by learning from past success and failure patterns to achieve up-to-date capabilities~\cite{zhang2026coevoskills}. As the number of skills continues to grow, scaling manual refinement becomes increasingly impractical. This hurdle motivates automatic strategies by leveraging skill execution records, including rich feedback signals and task-solving trajectories~\cite{ni2026trace2skill}. Such raw execution signals and trajectories are often noisy, mixing successful steps with irrelevant or failed ones. Therefore, reliable skill evolution requires capturing reusable execution patterns across multiple trajectories rather than individual behaviors~\cite{zhang2026coevoskills}. Moreover, growing skill libraries likely introduce conflicting contents, leading to redundant storage, excessive token consumption, and poor generalizability~\cite{wang2026skillx, gao2026skillreducer, zhang2026memskill}. To address these hurdles, we outline evolution strategies especially along the source and granularity of the learning signal: \textbf{Execution feedback} operates on single-run step-level signals; \textbf{Trajectory distillation} operates on multi-run sequence-level patterns; \textbf{Compression and augmentation} operate on library-level structures; \textbf{Reinforcement learning} operates on task-level rewards. These paradigms are not mutually exclusive, but they represent the dominant design choices in the community. For example, execution feedback and trajectory distillation both learn from execution records but differ mainly in granularity: single-run, step-level signals versus multi-run patterns. Reinforcement learning methods likewise incorporate trajectory feedback and skill revision. We classify such methods under RL because task-level rewards ultimately drive their evolution. In this case, we also provide a decision-making guidance flowchart for evolution strategy selection (\autoref{fig:evolution_decision}). Furthermore, we include a comparison of skill evolution strategies, listing their execution models, tasks, performance, and open-source status to provide an overview and comparison (\autoref{tab:evolution_detailed}). These works contain different executor models, tasks, and evaluation protocols, so the reported numbers are not directly comparable. We further structure and analyze how the current evolution paradigms align with benchmarks, highlighting trade-offs and practical guidelines that motivate future research (\appref{supp_analysis}). 

%Rather than isolated techniques, these paradigms form a design spectrum: early-stage systems benefit from execution feedback and distillation to stabilize skill behaviors, while mature skill libraries require compression and RL-driven efforts to maintain efficiency and cross-task generalization (\appref{supp_analysis}). 

%These paradigms are not mutually exclusive, but they represent the dominant design choices in the community.
%We further structure and analyze how the current evolution paradigm align with benchmarks, highlighting trade-offs and practical guidelines that motivate future research (\appref{supp_analysis}).

%These four paradigms are not mutually exclusive, but they represent the dominant design choices in the community.
% These efforts emphasize automatic failure detection, knowledge distillation, compression, rewriting, and verification.

\input{tables/table_skill_evolution}

\paragraph{\catFeedbackTag}

The record of skill execution can reveal valuable feedback signals for skill improvement, including runtime errors, incorrect outputs, unmet task specifications, and execution paths. Inspired by human rewriting, an intuitive approach is to implement an automated loop that executes the existing skill, observes failure patterns from execution feedback, and then rewrites the skill to prevent such failures from recurring. The execution feedback can come from either clear signals~\cite{liu2026skillforge,zhang2026coevoskills,tian2026skills,si2026context,jin2025two,ju2026embodiskill} or implicit execution ones~\cite{yang2026autoskill, ma2026skillclaw,yang2026autoskill}, which are both crucial for guiding skill evolution.

Traceable signals of skill evolution can come from real engineering activities, revealing user intent, agent tool calling, and concrete error patterns. These signals are critical for automatically detecting, diagnosing, and correcting flawed skills. For instance, \textsc{SkillForge}~\cite{liu2026skillforge} creates new skills by detecting the discrepancies between execution and reference behaviors. In particular, SkillForge produces structured failure records to identify the systemic patterns, reducing the need for human rewriting and verification. To support multi-turn conversations, \textsc{CoEvoSkills}~\cite{zhang2026coevoskills} enables agents to reduce human–machine cognitive misalignment and produce self-evolved skills that outperform human-curated skills. Especially to address failed executed skills, it introduces a verifier that can provide direct feedback about root-cause analysis and revision suggestions. Accessing the rich environment feedback can further enhance skill reliability.~\textsc{EmbodiSkill}~\cite{ju2026embodiskill} leverages the agent execution feedback by interacting with the environment to produce a trajectory of actions, observations, and a final reward. Unlike relying on the real execution feedback, \textsc{Skills-Coach}~\cite{tian2026skills} executes skills on synthetic cases to achieve the evolution feedback. Skills-Coach produces several rewritten versions of the seed skill. The highest-scored rewritten skill serves as a successful signal to improve skill instructions, while the failure traces drive skill scripts to prevent failures. Similarly, \textsc{Ctx2Skill}~\cite{si2026context} learns from feedbacks by producing synthetic diagnostic questions from the reference document.

%Among its testings, failed cases drive skill evolution by rewriting the skill to accumulate knowledge, while the easier ones are routed to improve the difficulty of questions. Notably, leveraging both hardest and easiest questions to select intermediate skills strikes an effective balance between task challenge and skill performance.

Even without a clear execution signal, user's preferences across conversations, such as preferred tone, terminology, or writing conventions, remain valuable for improving skill evolution. We support that the interaction traces could contain rich signals with reusable knowledge. \textsc{AutoSkill}~\cite{yang2026autoskill} treats interactions from users as the main signal for skill evolution. Rather than relying solely on failure correction, it turns user preferences into explicit capabilities that personalize the agent’s behavior. Similarly, \textsc{SkillClaw}~\cite{ma2026skillclaw} leverages heterogeneous user experiences from key trajectories that reflect how different users interact with tools and workflows. 

Across above execution feedback studies, we identify that a structured failure mode becomes a meaningful design factor. Systems that well separate failure diagnosis from rewrite generation (\textsc{SkillForge}~\cite{liu2026skillforge}, \textsc{CoEvoSkills}~\cite{zhang2026coevoskills}) tend to report stronger cross-task results compared to systems operating on raw traces (\textsc{AutoSkill}~\cite{yang2026autoskill}, \textsc{SkillClaw}~\cite{ma2026skillclaw}), although head-to-head comparisons remain absent. It is evident that feedback signals are inherently bounded by execution environment diversity rather than deployment—a structural constraint worth explicit attention in design and evaluation of skills.

%Similarly, \textsc{AutoSkill}~\cite{yang2026autoskill} treats interactions from users as the main signal for skill evolution. Rather than relying solely on failure correction, it transforms user repeated preferences into explicit capabilities that personalize the agent’s behavior over time. 

%  Skill-coach: Finally, a separate module then compares the original and optimized skills on a held-out test set to decide whether to adopt the new version, without real user tasks or human labels. 
% Together, these works highlight a shift toward experience-driven learning, where the goal of skill evolution extends beyond fixing failures to mastering the nuanced expectations of users.
% AutoSkill: It improves skills from stable user requirements, such as following institutional writing conventions or avoiding overly technical wording. 
% SkillClaw: By sharing these improvements across the entire system, the preferences or methods of one user become available to all others. 

% The single execution trajectory often reveals only partial behavior. To mitigate this, r
\paragraph{\catDistillTag} Skill evolution via trajectory distillation gains momentum to improve skills through sequential memorization by capturing task-specific, reusable patterns. For instance,~\textsc{SPARK}~\cite{zhou2026evidence} explores online trajectory verification to distill strong skills from executable evidence. It introduces a key trajectory-level measure to assess skill performance using task-environment evidence rather than unverified prior plans. \textsc{Trace2Skill}~\cite{ni2026trace2skill} updates skills from multiple execution trajectories by generating targeted patches from success and failure cases, then merging redundant fixes into a single conflict-free skill file. \textsc{Memento-Skills}~\cite{zhou2026memento} introduces a read-write reflective loop for skill evolution: a router retrieves relevant skills for execution, and the agent updates them from execution feedback, enabling iterative refinement and long-term behavioral memory. To broaden the data modality, \textsc{XSkill}~\cite{jiang2026xskill} grounds skill learning in visual observations, capturing relations between visual states and decisions. It extracts task-level skills through visual summarization and action-level experiences through cross-rollout critiques of successes and failures, consolidating both into a unified skill bank through merging and refinement.

% Finally, a feedback loop from inference usage continuously refines the dual-stream knowledge base rather than leaving it frozen after initial extraction.

%  Trace2Skill: Skills evolved by a 35B model in this way were transferred to a 122B model, which lifted accuracy by up to 57.65\% on WikiTableQuestions. 

% \input{tables/table_skill_evolution}

\paragraph{\catCompAugTag}
%We therefore group recent works into two categories: (i) \emph{organizing} the library, detecting and merging duplicates, and (ii) \emph{compressing and composing} individual skills.

% SkillX builds a knowledge base with three skill levels: planning skills (high-level task decompositions), functional skills (reusable multi-step tool subroutines), and atomic skills (small tool-use patterns and constraints).

As skill libraries grow rapidly, skill overlap and conflict lead to redundant exploration and poor generalization~\cite{wang2026skillx,liang2026skillnet}. Therefore, skill compression and augmentation are increasingly important for reducing duplication, gaining complementary knowledge, and exploring reliable skills.~\textsc{SkillNet}~\cite{liang2026skillnet} enables the community to create, evaluate, and connect agent skills as sourced from GitHub repositories and office documents. The resulting skill-similarity relation graph becomes a key to indicate whether the existing skills should be reused or merged. However, SkillNet does not explicitly define skill categories to support its evolution. To address this, \textsc{SkillX}~\cite{wang2026skillx} builds a multi-category skill evolution derived from execution trajectories by merging similar skills, decomposing complex ones, and assessing their generalization. To broaden coverage, SkillX prioritizes underexplored or failure-prone tools and synthesizes novel tasks to acquire and validate new skills to improve skill quality, richness, and coverage.~\textsc{SkillFoundry}~\cite{shen2026skillfoundry} takes a knowledge-driven approach for skill augmentation, organizing a tree structure where each node tracks references and existing skills to prioritize underexplored branches. This structure mines heterogeneous scientific resources into executable skills. Alternatively, \textsc{SkillReducer}~\cite{gao2026skillreducer} reframes skill evolution as content cleanup, pruning overly long skill descriptions and reorganizing skills into actionable rules and references, preserving essential knowledge while reducing token overhead. As these skills move into deployment, a promising frontier is grounding compression and augmentation decisions in live signals such as retrieval frequency and runtime failure rates rather than the offline design alone. %SkillFoundry produces a library where 71.1\% of skills are novel relative to existing ones while improving downstream agent performance on scientific benchmarks.

% SkillReducer compresses 48\% of description tokens and 39\% of body tokens while improving agent performance. 
% enabling them to learn and decide whether to improve an existing skill or reuse an old one

% It organized the skill bank into task- and step-skills to update the policy with a clean signal in different scopes. The task skills are retrieved to provide high-level guidance, indicating the task's overall success. The step skills provide fine-grained advice.

\paragraph{\catRLTag} Reinforcement learning (RL)~\cite{li2026arise,zhou2026dare} has emerged as a principled approach for aligning LLM agents with task execution rewards and driving reliable skill evolution~\cite{wang2023voyager,wang2024agent,jin2025apeer}. However, standard RL only rewards single task per update, while the real value of a skill roots in its reusability across tasks~\cite{tu2026dynamic}. To achieve stable rewards,~\textsc{D2Skill}~\cite{tu2026dynamic} leverages the multiple-rollout strategy in GRPO, enabling the policy agent to generate highly reusable skills. For each task, the LLM agent retrieves the most relevant skills and runs them twice (with the retrieved skills and without). The resulting success-rate gap between the two rollouts yields more stable rewards and improved skill reusability. Also, \textsc{SkillRL}~\cite{xia2026skillrl} leverages GRPO and exploits both success and failure signals, collecting trajectories across multiple rollouts to update the policy for skill retrieval and refinement by identifying failure patterns to drive targeted skill revision. However, these studies~\cite{tu2026dynamic, xia2026skillrl} treat skill retrieval, utilization, and evolution as separate components, risking conflicts during concurrent updates.~\textsc{Skill1}~\cite{shi2026skill1} overcomes this fragmentation via a unified co-evolution by training a single policy to jointly perform skill search, selection, task solving, and skill evolution within a single rollout. Overall, we observe that major efforts trace a clear trajectory in RL-based skill evolution. These approaches rely on task-level rewards that can conflate skill quality with agent capability, leaving open whether performance gains come from skill evolution or model improvement.

%% file: tables/table_skill_evolution.tex
% Requires in preamble:
%   \usepackage{booktabs}
%   \usepackage{tabularx}
%   \usepackage[table]{xcolor}
%   \usepackage{array}

% --- Category colors ---
\definecolor{catFeedback}{RGB}{255,229,204}   % light orange  — Execution Feedback
\definecolor{catDistill}{RGB}{217,234,211}    % light green   — Trajectory Distillation
\definecolor{catCompAug}{RGB}{207,226,243}    % light blue    — Compression and Augmentation
\definecolor{catRL}{RGB}{230,217,243}         % light purple  — Reinforcement Learning

% V1-END

\begin{table}[t!]
\centering
\footnotesize
\setlength{\tabcolsep}{4pt}
\renewcommand{\arraystretch}{1.25}
\renewcommand{\tabularxcolumn}[1]{>{\raggedright\arraybackslash}m{#1}}
\begin{tabularx}{\columnwidth}{@{}X@{}}
\toprule
\rowcolor{black!8}\textbf{Evolution Strategy}\\
\midrule
\rowcolor{catFeedback}\textbf{\faSync~Execution Feedback}\\
\addlinespace[1pt]
\textbf{SkillForge}~{\footnotesize\cite{liu2026skillforge}}\\
\textbf{CoEvoSkills}~{\footnotesize\cite{zhang2026coevoskills}}\\
\textbf{Skills-Coach}~{\footnotesize\cite{tian2026skills}}\\
\textbf{Ctx2Skill}~{\footnotesize\cite{si2026context}}\\
\textbf{AutoSkill}~{\footnotesize\cite{yang2026autoskill}}\\
\textbf{SkillClaw}~{\footnotesize\cite{ma2026skillclaw}}\\
\textbf{EmbodiSkill}~{\footnotesize\cite{ju2026embodiskill}}\\
\midrule
\rowcolor{catDistill}\textbf{\faRoute~Trajectory Distillation}\\
\addlinespace[1pt]
\textbf{SPARK}~\cite{zhou2026evidence} \\
\textbf{Trace2Skill}~{\footnotesize\cite{ni2026trace2skill}}\\
\textbf{Memento-Skills}~{\footnotesize\cite{zhou2026memento}}\\
\textbf{XSkill}~{\footnotesize\cite{jiang2026xskill}}\\

\midrule
\rowcolor{catCompAug}\textbf{\faCompress~Compression \& Augmentation}\\
\addlinespace[1pt]
\textbf{SkillNet}~{\footnotesize\cite{liang2026skillnet}}\\
\textbf{SkillX}~{\footnotesize\cite{wang2026skillx}}\\
\textbf{SkillReducer}~{\footnotesize\cite{gao2026skillreducer}}\\
\textbf{SkillFoundry}~{\footnotesize\cite{shen2026skillfoundry}}\\
\midrule
\rowcolor{catRL}\textbf{\faTrophy~Reinforcement Learning}\\
\addlinespace[1pt]
\textbf{D2Skill}~{\footnotesize\cite{tu2026dynamic}}\\
\textbf{SkillRL}~{\footnotesize\cite{xia2026skillrl}}\\
\textbf{SkillOS}~{\footnotesize\cite{ouyang2026skillos}}\\
\textbf{Skill1}~{\footnotesize\cite{shi2026skill1}}\\
\bottomrule
\end{tabularx}
\caption{Summary of skill evolution strategy in \autoref{sec:2.3}.
Categories are color-coded:
\catbox{catFeedback}~Execution Feedback,\;
\catbox{catDistill}~Trajectory Distillation,\;
\catbox{catCompAug}~Compression \& Augmentation,\;
\catbox{catRL}~Reinforcement Learning.}
\label{tab:evolution_categories}
\end{table}

%% file: sections/05_benchmarks.tex
\section{Skill-centric Evaluation and Benchmarks}
\label{sec:2.5}

Evaluation is important in the agent skill lifecycle because skills are continuously created, evolved, and shared among users. Without rigorous evaluation, it is difficult to fairly assess skill quality and safety. In principle, the skill evaluation should serve as: (1) validation of the comprehensive task performance; (2) comparison among skills in a fair environment; (3) safety auditing to detect harmful behaviors before skill deployments. We discuss skill-centric benchmarks to measure the realistic performance of agent skills. In \autoref{tab:skill_centric_benchmarks}, we group them into six major categories. The benchmarks we reviewed are not designed to identify which model performs best when equipped with a skill; rather, they ensure the base agent and tool environment are fixed and vary only the presence of the skill. We also include general-domain benchmarks that were not designed for skill evaluation yet are still applicable to assess agent skills (details in \appref{supp_general_domain_benchmarks}). 

\autoref{tab:skill_centric_benchmarks} reveals four structural gaps that warrant systematic investigation. First, utility and safety benchmarks collectively cover 11 professional domains and 581 auditable packages, whereas generation benchmarks span only 15 sub-domains with 20 core tasks. Second, no existing benchmark evaluates evolution longitudinally, i.e., tracking whether a skill improves across multiple rounds of feedback rather than measuring a single snapshot. Third, evaluation metrics are predominantly binary (pass/fail), which overlook operational factors such as token cost, latency, and error type. Fourth, a few details of benchmarks are not reported by the original benchmark articles, including annotation quality, verifier reliability, domain bias, license/accessibility, and running cost. These gaps should guide future research as much as the benchmarks themselves.

\input{tables/tab_07_dynamic_benchmarks}

\paragraph{\catUtilityTag}
Skill utility remains the primary criterion for assessing how skills improve an agent's task completion performance~\cite{li2025skillflow,zhang2026coevoskills,gao2026skillreducer}.~\textsc{SkillsBench}~\cite{li2026skillsbench} covers 86 hand-built tasks across 11 professional domains with deterministic verifiers. Across 7{,}308 trajectories and 7 agent-model configurations, curated skills raise the average 16\% pass rate with effects ranging from $+4.5$ pp in software engineering to $+51.9$ pp in healthcare. SkillsBench has since become a source for following benchmarks that extend the evaluation scope. For instance, SkillTester~\cite{wang2026skilltester} jointly measures utility and safety and SkillRouter~\cite{zheng2026skillrouter} evaluates skill retrieval and reranking. Alternatively, \textsc{SkillCraft}~\cite{chen2026skillcraft} focuses on identifying whether agents can reuse their own tool compositions across tasks. In particular, SkillCraft introduces long-horizon tool-use tasks whose difficulty is scaled along two axes: a quantitative axis that increases the number of agent-process entries, and a structural axis that lengthens the tool-call chain by composing subtasks into deeper workflows. Under its protocol, a compelling trait is that an agent can package the successful tool sequence as a reusable skill for later tasks. To date, the tool-calling cost, latency, and reasoning quality remain underexplored factors that future benchmarks should explicitly address.

\paragraph{\catGenTag}
The skill generation benchmarks provide a quantitative evaluation of skill quality, which is increasingly critical as manual quality assessment becomes challenging nowadays. As hundreds of skills are produced, early identification of low-quality skills could reduce token consumption. \textsc{SkillLearnBench}~\cite{zhong2026skilllearnbench} contains 20 skill-dependent tasks across 15 sub-domains, each with multiple variants that have the same task structure with different input values, enabling evaluation of skill reuse beyond one-shot success. SkillLearnBench indicates that external execution feedback is essential for skill improvement by preventing error accumulation. To better evaluate skill generation, research efforts should assess skill document quality, adherence between execution and prescribed steps, and task completion success.

\paragraph{\catRetrievalTag}
%As skill libraries grow rapidly, simply listing every skill in context becomes infeasible, and a new class of benchmarks is developed to evaluate the retrieval and routing performance of the methods for skill usage.

Skill routing and retrieval benchmark focuses on evaluating the effectiveness and accuracy of skill usage. Effective skill-retrieval methods must distinguish near-duplicate skills, while a well-designed routing should coordinate the appropriate skills for each task. \textsc{SkillRouter}~\cite{zheng2026skillrouter} contains roughly 80K skills and 75 expert-verified queries. SkillRouter demonstrates that only skill names and descriptions could result in a 31-44\% drop in routing accuracy compared to using the full skill body. \textsc{SRA-Bench}~\cite{su2026skill} decomposes skill augmentation into three separate stages: retrieval, incorporation, and application. It mixes 636 manually written gold skills into a 26{,}262-skill web-collected corpus and pairs them with 5{,}400 capability-intensive instances drawn from TheoremQA, LogicBench, ToolQA, MedCalc-Bench, CHAMP, and BigCodeBench. \textsc{AgentSkillOS Benchmark}~\cite{li2026organizing} shifts the focus of evaluation to the orchestration that allows multiple skills to work together on a task. It constructs 30 artifact-rich tasks across data computation, document creation, visual design, and web interaction. Evaluating at ecosystem scales from 200 to 200K skills, this orchestration substantially outperforms single-skill approaches when using the same skill set.

\paragraph{\catSafetyTag}
Skill security threats become increasingly systemic as malicious skills can compromise user data, hijack execution flows, and silently degrade agent behavior.~\textsc{SkillTester}~\cite{wang2026skilltester} jointly assesses utility and safety by executing candidate skills, reporting a composite security score alongside a three-level status label (e.g., safe, warning, or unsafe) that allows users to weigh performance gains against known risks. \textsc{SkillGuardBench}~\cite{lv2026structured} treats each skill as a multi-file package labeled as benign, suspicious, or malicious. The benchmark is built upon 581 packages across five evaluation views, with risk samples covering three recurring attack patterns: hidden override, disguised transfer, and remote bootstrap. In addition to auditing skills, \textsc{SKILL-INJECT}~\cite{schmotz2026skill} shifts the threat model towards runtime vulnerability, testing whether agents will execute malicious instructions embedded inside otherwise legitimate skills. Although safety auditing benchmarks have advanced, they treat safety as a one-time gate, leaving post-installation skill behavior across evolving libraries largely unexamined.

% Every risk seed is paired with semantics-preserving rewrites so that a dedicated attack-exact consistency metric can measure whether a guardrail maintains the same label under surface-form changes. %Together, SkillTester and SkillGuardBench frame skill auditing as a distinct evaluation problem from utility benchmarking, complementing SkillsBench-style pass-rate measurement with safety and rewrite-robustness signals.

% It found that 39 of the 49 evaluated skills yield zero pass-rate improvement and the average gain is only $+1.2$\%. 
\paragraph{\catSWETag}
The growth of skills in software engineering demands more rigorous benchmarking. Current efforts are strongly tied to public repositories, leaving the real-world engineering use unexplored.~\textsc{SWE-Skills-Bench}~\cite{han2026swe} takes a meaningful step toward closing this gap by pairing 49 publicly available software-engineering skills with GitHub repositories and evaluating them across 565 automated task instances. Its reliance on publicly curated skills and fixed-commit repositories allows good reproducibility; however, it inherently omits proprietary workflows, legacy codebases, and continuously evolving engineering practices. Future benchmarks should shift towards this trend by incorporating industry-partnered task suites, dynamic repository states, and evaluation metrics that capture maintenance overhead, error recovery rate, and token efficiency. Meanwhile, stratifying skills by authorship expertise offers an opportunity to advance the field from implementation-level evaluation toward architecture-level understanding.

%Each task is described by a requirement document whose acceptance criteria are converted into automated tests, yielding about 565 task instances across six subdomains. The underlying execution environment is \textsc{SWE-bench} and its human-validated \textsc{SWE-bench Verified} subset~\cite{jimenez2024swe}, which provides real GitHub issues requiring multi-file edits. 

% , yet it is unclear whether company-authored skills would behave similarly. 
% , showing that rapid skill adoption in software engineering has outrun real utility.

\paragraph{\catWildTag}
Evaluating agent skills in real-world environments is difficult to standardize, but it is essential because only dynamic, open-ended settings can reveal true deployment readiness of skills.~\textsc{WildClawBench}~\cite{ding2026wildclawbench} drops agents into a live personal-assistant environment and runs 60 hand-built original tasks across productivity flow, code intelligence, social interaction, and safety alignment. Users can submit results from their own customized agents, turning the benchmark into a community-driven testbed for skill ecosystems. Similarly, SkillForge~\cite{liu2026skillforge} introduces a benchmark of five real-world cloud technical-support scenarios spanning 1{,}883 tickets and 3{,}737 tasks. While both benchmarks advance the real-world evaluation, their closed-environment designs restrict external reproducibility and cross-method comparison. Moving forward, the field would benefit from open skill execution environments with standardized task interfaces, paired with auditable abilities of skill library updates.

%% file: tables/tab_07_dynamic_benchmarks.tex
\makeatletter
\@ifundefined{LT@col@L}{%
  \newcolumntype{L}[1]{>{\raggedright\arraybackslash}p{#1}}%
}{}
\@ifundefined{LT@col@C}{%
  \newcolumntype{C}[1]{>{\centering\arraybackslash}p{#1}}%
}{}
\makeatother

% V1-END

\begin{table*}[t!]
\centering
\small
\setlength{\tabcolsep}{4pt}
\renewcommand{\arraystretch}{1.25}
%% Make tabularx X column vertically centered (default is top-aligned p{}).
\renewcommand{\tabularxcolumn}[1]{>{\raggedright\arraybackslash}m{#1}}
\begin{tabularx}{\textwidth}{@{}>{\raggedright\arraybackslash}m{2.6cm} >{\raggedright\arraybackslash}m{3.4cm} X@{}}
\toprule
\rowcolor{black!8}\thead{\textbf{Benchmark}} & \thead{\textbf{Scale (Total)}} & \thead{\textbf{Task Composition}} \\
\midrule
%% --- Utility -----------------------------------------------------------
\rowcolor{catUtility}\multicolumn{3}{@{}l}{\textbf{\faWrench~Utility}}\\
% %\addlinespace[1pt]
\textbf{SkillsBench}~{\footnotesize\cite{li2026skillsbench}}
  & \makecell[l]{86 tasks\\(7{,}308 trajectories;\\7 agent--model configs)}
  & 11 professional domains: healthcare, manufacturing, cybersecurity, natural science, energy, office \& white collar, finance, media \& content production, robotics, mathematics, software engineering \\
% %\addlinespace[1pt]
\textbf{SkillCraft}~{\footnotesize\cite{chen2026skillcraft}}
  & 126 tasks
  & Long-horizon compositional tool-use tasks scaled by item count and tool-call chain depth; agents cache successful tool sequences as a persistent skill library \\
\midrule
%% --- Generation --------------------------------------------------------
\rowcolor{catGen}\multicolumn{3}{@{}l}{\textbf{\faSeedling~Generation}}\\
%\addlinespace[1pt]
\textbf{SkillLearn\-Bench}~{\footnotesize\cite{zhong2026skilllearnbench}}
  & \makecell[l]{20 tasks,\\100 instances}
  & 6 categories, 15 sub-domains (software engineering, information retrieval, productivity tools, data \& analytics, content \& creative, utilities); 3-level evaluation (skill quality, trajectory alignment, task outcome) \\
\midrule
%% --- Retrieval & Routing ----------------------------------------------
\rowcolor{catRetrieval}\multicolumn{3}{@{}l}{\textbf{\faSearch~Retrieval \& Routing}}\\
%\addlinespace[1pt]
\textbf{SRA-Bench}~{\footnotesize\cite{su2026skill}}
  & \makecell[l]{5{,}400 instances\\(636 gold skills in \\26{,}262 skill corpus)}
  & 6 source datasets (TheoremQA, LogicBench, ToolQA, MedCalc-Bench, CHAMP, BigCodeBench); decomposed evaluation of skill retrieval, incorporation, and application \\
%\addlinespace[1pt]
\textbf{SkillRouter}~\\{\footnotesize\cite{zheng2026skillrouter}}
  & \makecell[l]{75 core queries\\($\sim$80K candidate skills)}
  & SkillsBench-derived routing benchmark; compares metadata-only vs.\ full-body retrieval and reranking \\
%\addlinespace[1pt]
\textbf{AgentSkillOS}~{\footnotesize\cite{li2026organizing}}
  & \makecell[l]{30 tasks\\(200 to 200{,}000 skills)}
  & Data computation, document creation, motion video, visual design, web interaction \\
\midrule
%% --- Safety & Security -------------------------------------------------
\rowcolor{catSafety}\multicolumn{3}{@{}l}{\textbf{\faLock~Safety \& Security}}\\
%\addlinespace[1pt]
\textbf{SkillTester}~{\footnotesize\cite{wang2026skilltester}}
  & Per-skill (variable)
  & 2 utility task groups (common functional, edge functional) + 3 security probe groups (abnormal behavior control, permission boundary, sensitive data protection); outputs utility score, security score, and 3-level security status label \\
%\addlinespace[1pt]
\textbf{SkillGuard\-Bench}~{\footnotesize\cite{lv2026structured}}
  & \makecell[l]{581 packages\\(327 core\\+ 254 public-ecosystem);\\5 evaluation views\\(254--404 packages each)}
  & Package-level (\texttt{SKILL.md} + scripts + references + repo context) auditing; 3-way labels (benign / suspicious / malicious) covering hidden override, disguised transfer, remote bootstrap; semantics-preserving rewrites for attack-exact-consistency \\
\textbf{SKILL-INJECT}~{\footnotesize\cite{schmotz2026skill}}
  & \makecell[l]{23 skills, \\202 injection-task pairs}
  & 8 categories: data exfiltration, data destruction, DoS, ransomware, phishing, backdoors, bias manipulation, poisoning\\
\midrule
%% --- Software Engineering ---------------------------------------------
\rowcolor{catSWE}\multicolumn{3}{@{}l}{\textbf{\faLaptopCode~Software Engineering}}\\
%\addlinespace[1pt]
\textbf{SWE-Skills-Bench}~{\footnotesize\cite{han2026swe}}
  & \makecell[l]{$\sim$565 task instances\\(49 public SWE skills)}
  & 6 SWE subdomains over authentic GitHub repos pinned at fixed commits; requirement docs with deterministic execution-based acceptance criteria; paired with/without-skill evaluation \\
\midrule
%% --- Real-world environment -------------------------------------------
\rowcolor{catWild}\multicolumn{3}{@{}l}{\textbf{\faGlobe~Real-world Environment}}\\
%\addlinespace[1pt]
\textbf{WildClaw\-Bench}~{\footnotesize\cite{ding2026wildclawbench}}
  & 60 hand-built tasks
  & 6 categories in a live OpenClaw environment: productivity flow, code intelligence, social interaction, search \& retrieval, creative synthesis, safety alignment; Docker-isolated grading injected post-execution \\
%\addlinespace[1pt]
\textbf{SkillForge benchmark}~{\footnotesize\cite{liu2026skillforge}}
  & \makecell[l]{3{,}737 tasks\\(1{,}883 tickets)}
  & Five real-world cloud technical-support scenarios. \\
\bottomrule
\end{tabularx}
\caption{Skill-centric dynamic benchmarks. Categories are color-coded:
\catbox{catUtility}~Utility,\;
\catbox{catGen}~Generation,\;
\catbox{catRetrieval}~Retrieval \& Routing,\;
\catbox{catSafety}~Safety \& Security,\;
\catbox{catSWE}~Software Engineering,\;
\catbox{catWild}~Real-world environment.}
\label{tab:skill_centric_benchmarks}
\end{table*}

%% file: sections/06_discussion.tex
\section{Reflection and Future Directions}
\label{sec:discussion}

% TODO: Industry challenges, connection to Harness Agent Engineering, Security

Evaluation and evolution are becoming cornerstones of trustworthy agent skills. A central question is how to transform heterogeneous experiences, including human-written instructions, execution traces, user feedback, tool calls, and multimodal observations, into reliable and verifiable knowledge. Despite recent progress, open challenges persist in the handling of multimodal skills, effective use of trajectory data, and skill security, all of which demand systematic research efforts into robust evolution and evaluation frameworks.

% \paragraph{Multimodal skills development.}
% % \label{subsec:multimodal_skill}

Major skill frameworks use text-centered procedural packages that work for language, code, document, and API tasks, yet are ill-suited for agents operating in multimodal environments. Rich multimodal examples include desktop interfaces, web pages, embodied simulators, robotics, medical images, and visually grounded scientific workflows~\cite{zhou2025led}. In these scenarios, the right agent action is not from a textual goal alone; it also highly depends on the visual state, spatial layout, object configuration, interface affordances, and modality-specific constraints~\cite{jiang2026xskill}. A multimodal skill should identify the target visual elements, interpret the current state, and map it back to the content described in the skill. Skills that encode visual cues about the target retain success better than text-only procedural skills when a task is moved across visually distinct but functionally equivalent environments (e.g., alternative UI themes or robot embodiments), yielding a smaller success-rate drop under visual shift. XSkill~\cite{jiang2026xskill} offers insights into action-level, context-specific tool selection, and its skills capture task-level procedural knowledge for planning. Skill retrieval is driven by visual observations, which are revealed in past trajectories. In physical embodied settings, skill evolution could be driven by the execution feedback from environmental interactions, which reflects the agent's actions, observations, and rewards~\cite {ju2026embodiskill}. Yet the design, evaluation, and sharing of cross-modality skills that enable agents to act across diverse real-world sensory inputs remain largely underexplored.

% \paragraph{Trajectory data utility.}
% \label{subsec:trajectory_skills}
 
Trajectory data records~\cite{zhou2026evidence,ni2026trace2skill} are crucial for broadening agent skill utility by revealing intermediate reasoning, tool choices, recovery attempts, and failure modes. However, raw trajectories are often redundant and noisy. A successful trace can contain irrelevant steps, while a failed one may still hold useful local decisions~\cite{ni2026trace2skill}. To build robust agent skills, two emerging designs could make knowledge distillation from trajectory data more effective. First, distillation operates over batches of trajectories rather than single runs, since comparing many traces against each other is what isolates reusable patterns from task-specific noise~\cite{ni2026trace2skill}. Second, distillation can be continuous rather than one-shot, with skills serving as an evolving memory that is iteratively updated to reflect environmental changes~\cite {zhou2026evidence,zhou2026memento}. To unlock the full value of trajectory data, explicit curation along quality, diversity, and difficulty dimensions is essential to the skill evolution.
% Also, and through verifier-guided validation and provenance tracking afterward, so that each skill retains a lightweight link to the evidence behind it.

% \paragraph{Safety Concerns of Agent Skills}

Agent skills introduce non-trivial security risks that warrant systematic evaluation~\cite{wang2026skilltester,schmotz2026skill}. Malicious skills could manipulate LLMs to leak sensitive data, execute unauthorized commands, or produce harmful decisions. We identify three principal sources of skill poisoning that should be detected early and strictly avoided during skill evolution. First, direct instruction poisoning embeds harmful instructions into the skill, causing unsafe behavior during execution. Second, prompt injection occurs when a benign skill pulls content from untrusted external sources that carry malicious instructions. Third, uncontrolled skill self-evolution can silently strip existing safety constraints through unregulated updates. Further, skills distilled from execution trajectories risk unintentional privacy leakage through the skill body or outputs. Such risks are particularly acute in safety-critical domains such as healthcare and finance, where human oversight should be mandatory before skill deployment. The evolution mechanisms surveyed in (\autoref{sec:2.3}) can themselves amplify these risks. Methods that distill skills from user interactions, e.g., AutoSkill and SkillClaw, may mistakenly absorb injected malicious patterns, treating them as legitimate user preferences. Meanwhile, uncontrolled self-evolution loops, e.g., Memento-Skills and Skill1, may quietly strip away safety constraints as skills are repeatedly updated. Current auditing benchmarks cannot detect these risks because they assess a skill only once and overlook how danger accumulates across successive evolution rounds. This gap suggests that safety auditing must become longitudinal, tracking poisoning as a property of the full evolution trajectory rather than a single snapshot. To avoid such skill poisoning, we believe that a robust defense requires multi-layered approaches: a) establishing public reputation systems that track skill authorship, b) enforcing fine-grained permission boundaries on skill scripts, c) requiring explicit user confirmation before skills trigger sensitive actions suppresses prompt-injection attacks more than direct instruction poisoning, since confirmation intercepts externally triggered actions but not intent already embedded in the skill body.

Current evolution and evaluation strategies are largely treated as sequential, decoupled stages: a skill is evaluated, judged, and then evolved, after which evaluation restarts from scratch. This pipeline assumption is increasingly untenable at scale since the cost of re-evaluating an entire library after each evolution cycle becomes prohibitive. Two emerging approaches begin to close this gap from a joint learning perspective. SkillOS~\cite{ouyang2026skillos} builds an experience-driven RL recipe that pairs a frozen executor with a trainable skill curator using grouped task streams. Earlier trajectories update the skill repository, while later related tasks immediately evaluate those updates, leading to the effective evaluation via a structural component of the training loop rather than an external judge. Meanwhile, SkillsVote~\cite{liu2026skillsvote} approaches the same problem through a lifecycle governance, profiling a million-scale corpus for quality and verifiability. It attributes post-execution outcomes to skill use versus environment signals, and admits only successful discoveries through evidence-gated updates, showing that governed skill libraries can improve frozen agents without any model updates.

Realizing a unified framework that operates over living skill libraries more broadly demands meaningful progress on three key fronts. First, skill libraries must be engineered with explicit versioning and dependency graphs, such that a localized update to one skill can be automatically tested for downstream effects as evidenced in SkillsVote~\cite{liu2026skillsvote} and SkillX~\cite{jiang2026xskill}. Second, evaluation signals should move beyond binary pass/fail toward composite rewards that capture latency, token cost, and generalizability. SkillOS's composite reward design and SkillReducer's token-efficiency setting begin to address this challenge~\cite{gao2026skillreducer}. Third, skill curators must generalize across executor backbones and task domains. SkillOS~\cite{ouyang2026skillos} shows empirically across multi-turn and single-turn reasoning settings, yet no existing benchmark treats cross-domain curator generalization as an evaluation target. Together, these directions suggest redesigning skill libraries from static repositories into a more living, monitored infrastructure, where evolution and evaluation can be two faces of the shared continuous learning process~\cite{gao2025survey}.

We organize the above directions along a near-to-long-term spectrum to help readers prioritize their research directions. In the near term, most amount to engineering problems that extend existing pipelines, such as enriching evaluation rewards to capture latency, adding explicit versioning and dependency graphs for downstream testing, and deploying practical anti-poisoning safeguards. Moving toward the medium term, challenges such as coupling evolution and evaluation into a shared learning loop and continuously curating noisy trajectories require the development of new algorithmic frameworks. In the long term, cross-modality skill representation and provable skill security remain foundational, demanding deeper advances in how agents ground, share, and verify knowledge.

%% file: sections/07_conclusion.tex
\section{Conclusion}
\label{sec:conclusion}
In the era of rapid skills growth, the ability to continuously evolve and rigorously evaluate skills becomes vital to trustworthy agent deployment. We introduce a four-class taxonomy of skill evolution strategies, revealing that each class operates on distinct signal sources with performance trade-offs. Meanwhile, we analyze six skill-centric benchmark categories to thoroughly assess existing skills for public use. We identify structural gaps in longitudinal evaluation, retrieval coverage, and metric richness that should guide the next generation of benchmarks. Finally, future research should treat skill ecosystems as evolving infrastructure, where continual evaluation and evolution are central for reliable use, dependency control, and real-world agent deployment.

% Must include limitations 
\input{sections/08_limitation}

% Long Papers
% Long papers must describe substantial, original, completed and unpublished work. Wherever appropriate, concrete evaluation and analysis should be included. Long papers may consist of:

% up to eight (8) pages of content
% unlimited extra space after the conclusion for limitations (required, see below) and optional section on ethical considerations (we recommend it to be titled ‘Ethical considerations’)
% plus unlimited pages of references
% Submissions that exceed the length requirements, or are missing a limitations section, will be desk rejected.

%% file: sections/08_limitation.tex
\section*{Limitations}
\label{sec:limitations}
 
This survey provides an overview of agent skill evolution and evaluation with several limitations. First, given the rapid development of agent skill research, some recent methods or benchmarks may not be fully covered. Second, the reviewed systems are evaluated on different benchmarks and base models; we report findings as described rather than providing a unified empirical comparison, which is better addressed by a dedicated benchmarking study. Finally, we draw primarily on published papers and public repositories, and may therefore understate industrial practices where implementation details and proprietary skills remain undisclosed.

\section*{Ethical Considerations}
AI assistant was used to refine the appendix table (i.e., \appref{tab:general_domain}). All technical contents and final manuscript materials were reviewed and verified by the authors.

\section*{Acknowledgments}
We would like to thank the reviewers, area chairs, and program chairs for their evaluation and feedback. This research was supported by research grants to Dimitris N. Metaxas through NSF grants 2212302, 2212301, 2212303, AFOSR 23RT0630, and NIH 2R01HL127661.

%% file: sections/09_supp_skill_usage.tex
\section{Skill Usage}
\label{supp_skill_usage}
Skills are indexed by their name and description for rapid retrieval, while the full content will be loaded only upon selection. Yet SkillRouter~\cite{zheng2026skillrouter} finds that skill names and descriptions alone are inaccurate for skill selection at scale. Instead, SkillRouter adopts a retriever and a reranker to determine candidate skills by using full skill content. To reduce the cost of skill retrieval, SkillFlow~\cite{li2025skillflow} avoids repeating skill retrieval by first identifying the missing skill required to solve the task, then querying an external agent for the successfully executed skill, and saving it locally for future use. Unlike retrieval, effective routing enables the LLM agent to coordinate the appropriate skill to a specific task. SkillOrchestra~\cite{wang2026skillorchestra} compares success and failure trajectories to detect missing capabilities, which are summarized as new skills to build a skill handbook that can be consulted to identify required skills and route the task to the appropriate agent. Skill management organizes and updates a collection of skills, including removing redundant skills, pruning low-quality ones, refining skills as an up-to-date version, and controlling the size of skill library. For example, AgentSkillOS~\cite{li2026organizing} organizes skills into a capability tree, where the tree nodes are determined by skill categories and store the skill content. To keep the tree manageable, only the top-ranking skills are retained. AgentSkillOS traverses the tree to retrieve skills and caches successful orchestration plans for reuse. Similarly, SSL (Scheduling-Structural-Logical)~\cite{liang2026skill} converts the original skill text into a graph to better organize its content, including skill interface signals, operational stages, and individual actions. Different from implicit RL-trained routing policies, an interpretable handbook avoids routing collapse, transfers across orchestrator backbones without retraining, and achieves up to 22.5\% accuracy gains at 700× lower learning cost than RL methods.

%% file: sections/09_supp_general_domain.tex
\section{General-domain Benchmarks}
\label{supp_general_domain_benchmarks}
Although general-domain benchmarks were not originally designed for skill evaluation, they can be readily adapted to assess the performance of agents that learn and apply skills as below. 

% ─── Category 2: Interactive Agent Environments ──────────────────────────────
\paragraph{Interactive Agent Environments.}
Interactive agent environments evaluate an agent's ability to perceive the environment, plan multi-step actions, and execute long-horizon tasks under current observation states. They are adaptable, using guidance from skills to assess an agent's performance and determine skill quality. \textsc{ALFWorld}~\cite{shridhar2020alfworld} aligns text-based interactive household task completion with embodied ALFRED goals and TextWorld games, requiring agents to navigate rooms, manipulate objects, and follow natural-language instructions. \textsc{WebShop}~\cite{yao2022webshop} simulates online shopping over 1.18M real product listings with 12{,}087 crowd-sourced instructions, evaluating product search, attribute comparison, and goal-directed purchasing. \textsc{ScienceWorld}~\cite{wang2022scienceworld} provides an interactive text environment at the level of an elementary-school science curriculum, with 30 benchmark tasks (and 7{,}200 parametric variations) spanning thermodynamics, electrical circuits, chemistry, and biological processes. \textsc{WebArena}~\cite{zhou2024webarena} offers 812 realistic long-horizon web-based tasks requiring multi-step browser interaction across four real-world web applications (e-commerce, social forums, collaborative development, content management). \textsc{AgentBench}~\cite{liu2024agentbench} consolidates eight distinct interactive environments (e.g., operating system, database, knowledge graph) into a unified evaluation framework for assessing LLM-as-Agent reasoning and decision-making. \textsc{AppWorld}~\cite{trivedi2024appworld} provides a controllable world of 9 day-to-day apps operable via 457 APIs, with 750 natural-agent tasks for benchmarking interactive coding agents over stateful application use.
 
% ─── Category 3: Code Generation and Software Engineering ────────────────────
\paragraph{Code Generation and Software Engineering.}
Code-generation and software-engineering benchmarks evaluate functional correctness, command-line proficiency, and efficiency of program synthesis. They are relevant to skill evaluation because reusable engineering knowledge, including algorithmic recipes, debugging routines, and build-system patterns, is naturally described in the skill packages, and performance on these benchmarks could indicate whether the existing skills are useful for improving code generation or solving software engineering tasks. \textsc{Terminal-Bench}~\cite{merrill2026terminal} provides 89 hard, realistic command-line tasks for evaluating raw agent harness capability through shell-based interactions and is used by SkillFlow~\cite{li2025skillflow}. \textsc{HumanEval}~\cite{chen2021evaluating} releases 164 hand-written Python programming problems with unit tests to measure functional correctness of code synthesized from docstrings. \textsc{MBPP}~\cite{austin2021program} (Mostly Basic Programming Problems) contains 974 entry-level Python tasks (374 train / 90 val / 500 test) crowd-sourced to cover programming fundamentals and standard-library usage. For code efficiency, \textsc{EffiBench-X}~\cite{qing2026effibench} is the first large-scale multi-language code efficiency benchmark covering Python, C++, Java, JavaScript, Ruby, and Go, and serves as the primary evaluation for EffiSkill~\cite{wang2026effiskill}. The \textsc{PIE dataset}~\cite{shypula2023learning} contains over 77K paired slow/fast competitive C++ programming submissions across 1{,}474 problems and is used by EffiSkill for offline mining of recurring slow-to-fast transformations.
 
% ─── Category 4: Mathematical Reasoning ──────────────────────────────────────
\paragraph{Mathematical Reasoning.}
Mathematical reasoning benchmarks evaluate multi-step symbolic and quantitative reasoning with verifiable answers. They are well-suited to skill evaluation as mathematical solution strategies are reusable across problems. \textsc{AMC/AIME}~\cite{aops_aime} are annual competition-level problem sets ($\approx$30 problems per year) from the American Mathematics Competitions and the American Invitational Mathematics Examination, used as in-distribution and out-of-distribution mathematical reasoning evaluations. \textsc{Omni-MATH}~\cite{gao2025omni} provides 4{,}428 Olympiad-level problems spanning 33 sub-domains, forming the out-of-distribution suite together with AMC/AIME for ARISE~\cite{li2026arise}, which trains on the \textsc{DeepScaleR}~\cite{luo2025deepscaler} dataset of approximately 40K math problem-answer pairs compiled from AIME, AMC, Omni-MATH, and Still. AIME is also used by Trace2Skill~\cite{ni2026trace2skill} as one of its math reasoning evaluation domains.

\input{tables/tab_06_static_benchmarks}
\paragraph{Question Answering and Knowledge-Intensive Tasks.}
Search-augmented QA benchmarks evaluate retrieval, multi-hop reasoning, and tool-augmented information seeking with verifiable answers. They are central to skill evaluation because skills that encapsulate query decomposition, evidence selection, and cross-document synthesis can be measured by improvements on questions whose answers cannot be retrieved in a single hop.
 
Single-hop benchmarks include \textsc{Natural Questions (NQ)}~\cite{kwiatkowski2019natural}, derived from 307{,}373 / 7{,}830 / 7{,}842 real anonymized Google search queries with Wikipedia answer annotations; \textsc{TriviaQA}~\cite{joshi2017triviaqa}, consisting of over 650K question-answer-evidence triples (95K author-written QA pairs); and \textsc{PopQA}~\cite{mallen2023not}, 14K QA pairs converted from Wikidata triples to probe long-tail entity knowledge. Multi-hop benchmarks include \textsc{HotpotQA}~\cite{yang2018hotpotqa} with 113K Wikipedia-based questions and supporting facts; \textsc{2WikiMultiHopQA}~\cite{ho2020constructing}, 192{,}606multi-hop questions combining structured and unstructured Wikipedia/Wikidata evidence; \textsc{MuSiQue}~\cite{trivedi2022musique}, 25K 2-4-hop questions systematically composed from connected single-hop pairs to enforce genuine multi-hop reasoning; and \textsc{Bamboogle}~\cite{press2023measuring}, 125 manually constructed 2-hop questions designed to expose the compositionality gap. Together these seven form the search-augmented suite used by SkillRL~\cite{xia2026skillrl}. \textsc{WikiTableQuestions}~\cite{pasupat2015compositional} contains 22{,}033 complex questions over 2{,}108 semi-structured Wikipedia tables requiring compositional semantic parsing, and is used by Trace2Skill~\cite{ni2026trace2skill} as an out-of-distribution evaluation that demonstrates cross-model skill transfer (up to $+57.65$\% absolute gain). For agent-style general-knowledge tasks, \textsc{GAIA (General AI Assistants)}~\cite{mialon2024gaia} provides 466 questions (166 validation / 300 sequestered test) requiring web search, tools, and multi-step reasoning, and \textsc{Humanity's Last Exam (HLE)}~\cite{phan2025humanity} contributes 2{,}500 expert-validated questions spanning mathematics, sciences, and humanities; both are used by Memento-Skills~\cite{zhou2026memento} to evaluate continual skill-based agent improvement, achieving $26.2$\% and $116.2$\% relative gains, respectively.
 
% ─── Category 6: Knowledge, Language, and Instruction Following ──────────────
\paragraph{Knowledge, Language, and Instruction-Following.}
Knowledge and instruction-following benchmarks evaluate the breadth of factual knowledge, open-ended dialogue, and adherence to user instructions. They could be extended to skill evaluation to assess the capability of skills in improving task analysis, response structuring, and tone/format adaptation. \textsc{MMLU}~\cite{hendrycks2020measuring} provides 15{,}908 multiple-choice questions across 57 subjects spanning STEM, humanities, social sciences, and professional domains. \textsc{AlpacaEval}~\cite{li2023alpacaeval} comprises 805 instruction prompts with GPT-4-based pairwise win-rate annotation against a reference model, \textsc{MT-Bench}~\cite{zheng2023judging} contains 80 multi-turn questions across 8 categories evaluated via LLM-as-judge, and \textsc{WildBench}~\cite{lin2025wildbench} consists of 1{,}024 challenging tasks carefully curated from over one million real WildChat user-chatbot conversation logs; collectively these form the instruction-following suite used by SkillOrchestra~\cite{wang2026skillorchestra}.
 
% ─── Category 7: Memory and Conversational Benchmarks ────────────────────────
\paragraph{Memory and Conversational Benchmarks.}
Memory-centric benchmarks evaluate whether agents can extract, consolidate, and recall information across long interaction histories or multi-turn dialogues. They are adapted to validate the performance of using skills for summarization, indexing, and retrieval over past experience, which enables agents to operate beyond a single context window. \textsc{LoCoMo}~\cite{maharana2024evaluating} provides very long dialogues with $\sim$300 turns spanning up to 35 sessions, accompanied by question-answering, summarization, and multimodal probes; \textsc{LongMemEval}~\cite{wu2024longmemeval} contributes 500 QA items over long chat histories under both synthetic and realistic settings; together with \textsc{HotpotQA}~\cite{yang2018hotpotqa} (multi-hop QA) and \textsc{ALFWorld}, these are used by MemSkill~\cite{zhang2026memskill} to evaluate learnable memory skills for extracting and consolidating information across long interaction histories. \textsc{StuLife}~\cite{cai2025building} simulates a student's holistic college journey across three core phases and ten sub-scenarios in a persistent, stateful campus environment (1{,}284 interdependent tasks spanning a full academic year), and is referenced by AutoSkill~\cite{yang2026autoskill} as a representative experience-driven lifelong learning benchmark for evaluating self-evolving agents.

\input{tables/analysis_table}

% ─── Category 8: Multimodal and Tool-Use Benchmarks ──────────────────────────
\paragraph{Multimodal and Tool-Use Benchmarks.}
Multimodal and tool-use benchmarks evaluate visual grounding, tool selection, and orchestration of external resources alongside language reasoning. They are promising to skill evaluation as many real-world skills are inherently multimodal or tool-mediated (e.g., reading a chart and querying an API), and their value cannot be captured by purely textual benchmarks. For multimodal continual learning, XSkill~\cite{jiang2026xskill} evaluates on five benchmarks spanning three domains. Visual agentic tool use is covered by \textsc{VisualToolBench}~\cite{guo2025beyond}, comprising 1{,}204 open-ended vision tasks (603 single-turn and 601 multi-turn) across five domains paired with detailed rubrics, and by \textsc{TIR-Bench}~\cite{li2025tir}, which evaluates agentic thinking-with-images reasoning across 13 diverse tasks requiring novel tool use for image processing and manipulation in chain-of-thought. Multimodal search and web browsing are covered by \textsc{MMSearch-Plus}~\cite{tao2025mmsearch}, a 311-task provenance-aware benchmark that requires extracting weak, localized visual cues and propagating them through iterative image-text retrieval, and \textsc{MMBrowseComp}~\cite{li2025mm}, a hand-crafted set of 224 questions specifically designed to assess multimodal retrieval and reasoning over image- and video-rich web content. A comprehensive multimodal-agent setting is provided by \textsc{AgentVista}~\cite{su2026agentvista}, which contains 209 tasks across 25 sub-domains in 7 categories requiring long-horizon hybrid tool use (web search, image search, page navigation, and code-based image processing). \textsc{SpreadsheetBench}~\cite{ma2024spreadsheetbench} provides 912 real-world spreadsheet manipulation tasks from online Excel forums with 2{,}729 test cases, used by Trace2Skill~\cite{ni2026trace2skill} for skill-deepening evaluation. For tool-calling and conversational control, \textsc{BFCL-v3}~\cite{patil2024gorilla} extends the Berkeley Function Calling Leaderboard toward 1{,}000 multi-turn function calling data, and \textsc{$\tau^2$-Bench}~\cite{barres2025tau} studies dual-control telecom-style dialogues with compositional simulated users and verifiable outcomes. \textsc{M$^3$-Bench}~\cite{zhou2025m} further covers multi-modal, multi-hop, and multi-threaded tool-use agents over 28 MCP servers exposing 231 tools.
 
% ─── Category 9: Embodied / Open-Ended Environments ──────────────────────────
\paragraph{Embodied / Open-Ended Environments.}
Embodied and open-ended environments evaluate exploration, lifelong learning, and the construction of compositional skill libraries in worlds without a fixed task distribution. Leveraging such benchmarks for skill evaluation could assess the capability of skills to support cumulative, transferable competence acquired through interaction with the environment. \textsc{MineDojo}~\cite{fan2022minedojo} is a Minecraft-based framework for open-ended embodied lifelong learning, providing a simulation suite with thousands of programmatic tasks and an internet-scale knowledge base of 730K+ YouTube videos with time-aligned transcripts, 6K+ free-form Wiki pages, and 340K+ Reddit posts with multimedia contents. It is the evaluation environment for Voyager~\cite{wang2023voyager}, which measures unique items obtained, distance traveled, and tech-tree milestone progression to evaluate compositional skill libraries built through automatic curricula.

%% file: tables/tab_06_static_benchmarks.tex
% Unified benchmark table — label \ref{tab:benchmarks_all}
% Requires in your preamble:
%   \usepackage[table]{xcolor}
%   \usepackage{tabularx}
%   \usepackage{booktabs}
 
% --- Category band colors (one per section in 05_benchmarks.tex) ---
\definecolor{catSkill}    {HTML}{FFE8C2}  % skill-centric (warm sand)
\definecolor{catInteract} {HTML}{D7E9FF}  % interactive envs (light blue)
\definecolor{catSWE}      {HTML}{D9F2D9}  % code gen / SWE (light green)
\definecolor{catMath}     {HTML}{FAD4E1}  % math (pink)
\definecolor{catQA}       {HTML}{FFE3CC}  % QA / knowledge-intensive (peach)
\definecolor{catKnow}     {HTML}{E5DAF5}  % knowledge / language / IF (lavender)
\definecolor{catMem}      {HTML}{D6F2EF}  % memory / conversational (mint)
\definecolor{catMM}       {HTML}{CFE9F1}  % multimodal & tool-use (light cyan)
\definecolor{catEmb}      {HTML}{F1F1F1}  % embodied / open-ended (light gray)
 
\begin{table*}[t!]
\centering
\scriptsize
\setlength{\tabcolsep}{3pt}
\renewcommand{\arraystretch}{1.15}
\begin{tabularx}{\textwidth}{>{\raggedright\arraybackslash}p{0.26\textwidth} >{\raggedright\arraybackslash}X >{\raggedright\arraybackslash}p{0.2\textwidth}}
\toprule
\textbf{Benchmark} & \textbf{Scale (train / val / test or total)} & \textbf{Category} \\
\midrule
 
% % ─── 1. Skill-Centric Dynamic Benchmarks ─────────────────────────────────────
% \rowcolor{catSkill}\multicolumn{3}{l}{\textit{Skill-Centric Dynamic Benchmarks}} \\
% \midrule
% \rowcolor{catSkill}SkillsBench~\cite{li2026skillsbench}            & 86 tasks across 11 professional domains & Skill-centric \\
% \rowcolor{catSkill}SkillRouter~\cite{zheng2026skillrouter} & $\sim$80K candidate skills; 75 expert-verified queries & Skill retrieval / routing \\
% \rowcolor{catSkill}SkillTester~\cite{wang2026skilltester}          & SkillsBench tasks + dedicated security probe suite & Skill safety / utility \\
% \rowcolor{catSkill}SWE-Skills-Bench~\cite{han2026swe}   & 49 public SWE skills $\times$ $\sim$565 tasks across 6 categories & Skill-centric (SWE) \\
% \rowcolor{catSkill}AgentSkillOS~\cite{li2026organizing} & 30 artifact-rich tasks across 5 categories; ecosystem scales from 200 to 200K skills & Skill orchestration \\
% \rowcolor{catSkill}WildClawBench~\cite{ma2026skillclaw}            & Real-world agentic tasks for collective skill evolution & Skill evolution \\
% \rowcolor{catSkill}SkillForge internal benchmark~\cite{liu2026skillforge} & 5 cloud-support scenarios; 1{,}883 tickets; 3{,}737 tasks & Skill-centric (cloud ops) \\
 
\midrule
 
% ─── 2. Interactive Agent Environments ───────────────────────────────────────
\rowcolor{catInteract}\multicolumn{3}{l}{\textit{Interactive Agent Environments}} \\
\midrule
\rowcolor{catInteract}ALFWorld~\cite{shridhar2020alfworld}         & 3{,}827 games across 6 task types (pick \& place, examine in light, clean/heat/cool \& place, pick two \& place) & Text / embodied \\
\rowcolor{catInteract}WebShop~\cite{yao2022webshop}                & 1.18M product listings; 12{,}087 instructions & Web / shopping \\
\rowcolor{catInteract}ScienceWorld~\cite{wang2022scienceworld}     & 30 task types; 7{,}200 parametric variations & Interactive science \\
\rowcolor{catInteract}WebArena~\cite{zhou2024webarena}             & 812 long-horizon web tasks across 4 web applications & Realistic web GUI \\
\rowcolor{catInteract}AgentBench~\cite{liu2024agentbench}          & 8 distinct environment types & Multi-env agent suite \\
\rowcolor{catInteract}AppWorld~\cite{trivedi2024appworld}          & 9 apps; 457 APIs; 750 autonomous agent tasks & App / coding control \\
 
\midrule
 
% ─── 3. Code Generation and Software Engineering ─────────────────────────────
\rowcolor{catSWE}\multicolumn{3}{l}{\textit{Code Generation and Software Engineering Benchmarks}} \\
\midrule
\rowcolor{catSWE}Terminal-Bench~\cite{merrill2026terminal}    & 89 terminal tasks & CLI agent \\
\rowcolor{catSWE}HumanEval~\cite{chen2021evaluating}                & 164 hand-written Python programming problems & Code correctness \\
\rowcolor{catSWE}MBPP~\cite{austin2021program}                        & 974 programming tasks & Code correctness \\
\rowcolor{catSWE}EffiBench-X~\cite{qing2026effibench}             & 623 problems across 6 programming languages & Code efficiency \\
\rowcolor{catSWE}PIE dataset~\cite{shypula2023learning}                 & $\sim$77K slow/fast C++ pairs over 1{,}474 problems & Code efficiency \\
 
\midrule
 
% ─── 4. Mathematical Reasoning ───────────────────────────────────────────────
\rowcolor{catMath}\multicolumn{3}{l}{\textit{Mathematical Reasoning Benchmarks}} \\
\midrule
\rowcolor{catMath}AMC / AIME~\cite{aops_aime}                        & Annual competition-level problem sets ($\approx$30 / year) & Math competition \\
\rowcolor{catMath}Omni-MATH~\cite{gao2025omni}                 & 4{,}428 Olympiad-level problems across 33 sub-domains & Math olympiad \\
\rowcolor{catMath}DeepScaleR~\cite{luo2025deepscaler}              & 40K problem-answer pairs & Math RL training \\
 
\midrule
 
% ─── 5. Question Answering and Knowledge-Intensive ───────────────────────────
\rowcolor{catQA}\multicolumn{3}{l}{\textit{Question Answering and Knowledge-Intensive Benchmarks}} \\
\midrule
\rowcolor{catQA}Natural Questions (NQ)~\cite{kwiatkowski2019natural}    & 307{,}373 train / 7{,}830 dev / 7{,}842 test queries & Single-hop QA \\
\rowcolor{catQA}TriviaQA~\cite{joshi2017triviaqa}                  & 650K question-answer-evidence triples & Single-hop QA \\
\rowcolor{catQA}PopQA~\cite{mallen2023not}                       & 14K QA pairs over long-tail Wikipedia entities & Single-hop QA \\
\rowcolor{catQA}HotpotQA~\cite{yang2018hotpotqa}                   & 113K Wikipedia QA pairs & Multi-hop QA \\
\rowcolor{catQA}2WikiMultiHopQA~\cite{ho2020constructing}               & 192{,}606 multi-hop QA pairs over Wikipedia & Multi-hop QA \\
\rowcolor{catQA}MuSiQue~\cite{trivedi2022musique}                  & 25K 2-4-hop QA pairs & Multi-hop QA \\
\rowcolor{catQA}Bamboogle~\cite{press2023measuring}                & 125 compositional 2-hop questions & Multi-hop QA \\
\rowcolor{catQA}WikiTableQuestions~\cite{pasupat2015compositional}        & 22{,}033 questions over 2{,}108 Wikipedia tables & Table QA \\
\rowcolor{catQA}GAIA~\cite{mialon2024gaia}                         & 466 questions & General assistant QA \\
\rowcolor{catQA} HLE~\cite{phan2025humanity}                            & 2{,}500 expert-validated questions & Expert exam \\
 
\midrule
 
% ─── 6. Knowledge, Language, and Instruction-Following ───────────────────────
\rowcolor{catKnow}\multicolumn{3}{l}{\textit{Knowledge, Language, and Instruction-Following Benchmarks}} \\
\midrule
\rowcolor{catKnow}MMLU~\cite{hendrycks2020measuring}                    & 15{,}908 multi-choice questions across 57 subjects & Knowledge, multi-task \\
\rowcolor{catKnow}AlpacaEval~\cite{li2023alpacaeval}               & 805 evaluation prompts & Instruction following \\
\rowcolor{catKnow}MT-Bench~\cite{zheng2023judging}                 & 80 multi-turn questions across 8 categories & Multi-turn dialog \\
\rowcolor{catKnow}WildBench~\cite{lin2025wildbench}                & 1{,}024 challenging real-user tasks from WildChat logs & Real-user instruction \\
 
\midrule
 
% ─── 7. Memory and Conversational ────────────────────────────────────────────
\rowcolor{catMem}\multicolumn{3}{l}{\textit{Memory and Conversational Benchmarks}} \\
\midrule
\rowcolor{catMem}LoCoMo~\cite{maharana2024evaluating}                  & 1{,}986 questions & Long-horizon memory \\
\rowcolor{catMem}LongMemEval~\cite{wu2024longmemeval}              & 500 QA items over long chat histories & Long-horizon memory \\
\rowcolor{catMem}StuLife~\cite{cai2025building}                     & 1{,}284 interdependent tasks across 3 phases / 10 sub-scenarios & Lifelong learning \\
 
\midrule
 
% ─── 8. Multimodal and Tool-Use ──────────────────────────────────────────────
\rowcolor{catMM}\multicolumn{3}{l}{\textit{Multimodal and Tool-Use Benchmarks}} \\
\midrule
\rowcolor{catMM}VisualToolBench~\cite{guo2025beyond}      & 1{,}204 open-ended vision tasks (603 single-turn / 601 multi-turn) across 5 domains & Multimodal tool use \\
\rowcolor{catMM}TIR-Bench~\cite{li2025tir}                    & 13 thinking-with-images tool-use tasks & Multimodal tool use \\
\rowcolor{catMM}MMSearch-Plus~\cite{tao2025mmsearch}           & 311 provenance-aware multimodal search tasks & Multimodal search \\
\rowcolor{catMM}MMBrowseComp~\cite{li2025mm}             & 224 hand-crafted multimodal browsing questions & Multimodal web browsing \\
\rowcolor{catMM}AgentVista~\cite{su2026agentvista}                 & 209 tasks across 25 sub-domains in 7 categories & Multimodal agent suite \\
\rowcolor{catMM}SpreadsheetBench~\cite{ma2024spreadsheetbench}     & 912 questions; 2{,}729 test cases (avg.\ 3 per instruction) & Spreadsheet manipulation \\
\rowcolor{catMM}BFCL-v3~\cite{patil2024gorilla}                    & 1{,}000 multi-turn function calling data & Tool calling \\
\rowcolor{catMM}$\tau^2$-Bench~\cite{barres2025tau}          &  2{,}285 tasks & Conversational agent \\
\rowcolor{catMM}M$^3$-Bench~\cite{zhou2025m}                 & 28 servers with 231 tools & Multimodal tool use \\
 
\midrule
 
% ─── 9. Embodied / Open-Ended ────────────────────────────────────────────────
\rowcolor{catEmb}\multicolumn{3}{l}{\textit{Embodied / Open-Ended Environments}} \\
\midrule
\rowcolor{catEmb}MineDojo / Minecraft~\cite{fan2022minedojo}       &  730K+ YouTube videos with time-aligned transcripts, 6K+ free-form Wiki pages, and 340K+ Reddit posts with multimedia contents & Embodied / open-ended \\
 
\bottomrule
\end{tabularx}
\caption[Skill evaluation benchmark summary.]{General-domain evaluation benchmarks. \textbf{Scale} reports the dataset size and the standard splits from the primary reference. \textbf{Category} corresponds to the type of tasks.}
\label{tab:general_domain}
\end{table*}

%% file: tables/analysis_table.tex
\begin{table*}[htbp]
  \centering
  % \caption{Cross-paradigm trade-offs and benchmark alignment for skill evolution strategies.}
  % \label{tab:paradigm_tradeoffs}
  \small
  \renewcommand{\arraystretch}{1.2}
  \setlength{\tabcolsep}{3.5pt}
  \begin{tabular}{@{}p{2.9cm}p{2.5cm}p{2.9cm}p{3.6cm}p{3.3cm}@{}}
    \toprule
    \textbf{Evolution Paradigm} & \textbf{Primary Signal Source} & \textbf{Strengths} & \textbf{Critical Trade-offs} & \textbf{Benchmark Coverage Gap} \\
    \midrule
    \textbf{Execution Feedback} & Runtime errors, verifier signals & High fidelity to real failures; easily auditable & Reactive; struggles with sparse or ambiguous signals & Lack of longitudinal tracking across feedback rounds \\
    \midrule
    \textbf{Trajectory Distillation} & Multi-run success/failure traces & Captures reusable reasoning patterns \& recovery paths & Noise accumulation; trajectory bloat inflates context windows & Few benchmarks measure distillation efficiency vs. token cost \\
    \midrule
    \textbf{Compression \& Augmentation} & Inter-skill similarity, knowledge graphs & Reduces redundancy; improves routing \& generalization & Risk of stripping safety constraints or domain nuance & Limited evaluation of post-composition fidelity \& conflict resolution \\
    \midrule
    \textbf{Reinforcement Learning} & Multi-task reward gaps, rollout comparisons & Optimizes reusability \& long-horizon orchestration & Reward hacking; high compute; unstable without curated baselines & Binary pass/fail metrics ignore composite utility/safety trade-offs \\
    \bottomrule
  \end{tabular}
\caption{Cross-paradigm trade-offs and benchmark alignment for skill evolution strategies.}
\label{tab:paradigm_tradeoffs}
\end{table*}

%% file: sections/10_analysis_comparison.tex
\section{Practical Guidelines for Skill Evolution System Design}
\label{supp_analysis}
% analysis. 

\input{Figures/Revision_workflow}

\autoref{tab:paradigm_tradeoffs} maps each evolution paradigm to its primary signal source, empirical strengths, critical trade-offs, and benchmark coverage gaps. Evolution paradigm performance varies due to distinct signal sources that we offer practical guidelines to advance this research frontier. 

\noindent~\textbf{Execution Feedback}: The feedback loop must better distinguish between identifying failures and generating the rewrite. Since these execution-feedback methods often excel at high-fidelity failure correction but their signals are sparse when execution environments are narrow or deterministic, and existing evaluations measure mostly single-round skill quality rather than tracking improvement across repeated feedback cycles.

\noindent~\textbf{Trajectory distillation}: The distillation operation is recommended to compare patterns across multiple runs to discover reusable knowledge. High-quality trajectories are required in advance and should be explicitly curated along quality and diversity before distillation rather than using all available traces indiscriminately.

\noindent~\textbf{Compression and augmentation}:  Compression operations targeting token efficiency can degrade skill utility by improperly removing task-critical procedural knowledge. So before compression, core executable steps should be annotated as a protected reference. After compression, the evolved skill should be executed against a held-out task set to confirm performance is preserved. 
 
 \noindent~\textbf{Reinforcement learning}: To verify that an RL approach genuinely improves skill quality rather than training the agent to bypass skills, we recommend a dual-rollout evaluation protocol: at regular training intervals, evaluating task performance both with and without skills, and treating the performance gap as the skill contribution signal. A shrinking training gap is an early sign that the agent is learning to solve tasks bypassing the skill library and should trigger a review of the reward design.

% We also provide a decision-making guidance flowchart for evolution strategy selection (\autoref{fig:evolution_decision}). Furthermore, we include a comparison of skill evolution strategies, listing their execution models, tasks, performance, and open-source status to provide an overview and comparison (\autoref{tab:evolution_detailed}).

\input{tables/tab_revision_comparison}

 %Consequently, robust skill evolution requires paradigm hybridization, i.e., coupling execution feedback with safety gating for high-stakes tasks, or pairing trajectory distillation with RL and token-aware compression for exploratory workflows. We support that future evaluations should shift more from snapshot utility to longitudinal, composite, and paradigm-aware assessment.

%% file: Figures/Revision_workflow.tex
% ============================================================
% Preamble needs:
%   \usepackage{tikz}
%   \usetikzlibrary{arrows.meta, positioning, shapes.geometric}
%   \usepackage[table]{xcolor}
%   \usepackage{fontawesome5}   % optional, for category icons
%
% Category colors (identical to table_skill_evolution.tex):
%   \definecolor{catFeedback}{RGB}{255,229,204}
%   \definecolor{catDistill}{RGB}{217,234,211}
%   \definecolor{catCompAug}{RGB}{207,226,243}
%   \definecolor{catRL}{RGB}{230,217,243}
% ============================================================

\begin{figure*}[t]
\centering
\footnotesize
\begin{tikzpicture}[
  node distance=6mm and 10mm,
  every node/.style={font=\footnotesize},
  qbox/.style={draw, rounded corners, align=left, fill=black!5,
               text width=6.2cm, inner sep=5pt},
  leaf/.style={draw, rounded corners, align=left, text width=6.4cm,
               inner sep=5pt},
  edgelbl/.style={font=\scriptsize\bfseries, fill=white, inner sep=1pt},
  arrow/.style={-{Stealth[length=2mm]}, thick},
]

% --- Question nodes (spine down the left) ---
\node[qbox] (start) {\textbf{START:} What signal exists?\\ \emph{(choose the dominant source)}};
\node[qbox, below=of start] (q1) {\textbf{Q1:} Is the library redundant or bloated?\\ \emph{(overlap, conflict, token cost)}};
\node[qbox, below=22mm of q1] (q2) {\textbf{Q2:} Can tasks be scored automatically?\\ \emph{(verifiable reward, many rollouts)}};
\node[qbox, below=26mm of q2] (q3) {\textbf{Q3:} One run, or many runs to compare?\\ \emph{(signal granularity)}};

% --- Leaf nodes (category outcomes, on the right) ---
\node[leaf, fill=catCompAug, right=of q1] (comp)
  {\textbf{\faCompress~Compression \& Augmentation} \emph{[library-level]}\\
   SkillNet~\cite{liang2026skillnet}, SkillX~\cite{wang2026skillx}, SkillReducer~\cite{gao2026skillreducer}, SkillFoundry~\cite{shen2026skillfoundry}};

\node[leaf, fill=catRL, right=of q2] (rl)
  {\textbf{\faTrophy~Reinforcement Learning} \emph{[task-level rewards]}\\
   D2Skill~\cite{tu2026dynamic}, SkillRL~\cite{xia2026skillrl}, SkillOS~\cite{ouyang2026skillos}, Skill1~\cite{shi2026skill1}};

\node[leaf, fill=catFeedback, right=22mm of q3, yshift=4mm] (feed)
  {\textbf{\faSync~Execution Feedback} \emph{[step-level]}\\
   SkillForge~\cite{liu2026skillforge}, CoEvoSkills~\cite{zhang2026coevoskills}, Skills-Coach~\cite{tian2026skills}, Ctx2Skill~\cite{si2026context}, AutoSkill~\cite{yang2026autoskill}, SkillClaw~\cite{ma2026skillclaw}, EmbodiSkill~\cite{ju2026embodiskill}};

\node[leaf, fill=catDistill, below=6mm of feed] (dist)
  {\textbf{\faRoute~Trajectory Distillation} \emph{[sequence-level]}\\
   SPARK~\cite{zhou2026evidence}, Trace2Skill~\cite{ni2026trace2skill}, Memento-Skills~\cite{zhou2026memento}, XSkill~\cite{jiang2026xskill}};

% --- Edges ---
\draw[arrow] (start) -- (q1);

\draw[arrow] (q1.east) -- node[edgelbl, above] {YES} (comp.west);
\draw[arrow] (q1) -- node[edgelbl, right] {NO} (q2);

\draw[arrow] (q2.east) -- node[edgelbl, above] {YES} (rl.west);
\draw[arrow] (q2) -- node[edgelbl, right] {NO} (q3);

\draw[arrow] (q3.east) -- node[edgelbl, above, pos=0.6] {SINGLE RUN} (feed.west);
\draw[arrow] (q3.east) -- node[edgelbl, below, pos=0.6] {MANY RUNS} (dist.west);

\end{tikzpicture}
\caption{Decision procedure for selecting a skill-evolution strategy, keyed to the dominant learning signal. Each terminal category lists the systems surveyed in \autoref{sec:2.3}.}
\label{fig:evolution_decision}
\end{figure*}

%% file: tables/tab_revision_comparison.tex
\begin{table*}[t]
\centering
\footnotesize
\setlength{\tabcolsep}{4pt}
\renewcommand{\arraystretch}{1.25}
\caption{Detailed comparison of the skill-evolution systems surveyed in \autoref{sec:2.3}.
Each result uses that paper's own metric and baseline and should be read within that paper's setting only. \textbf{OS} (open source): Yes is code released; No is no public code found.}
\label{tab:evolution_detailed}
\begin{tabularx}{\textwidth}{@{}
  >{\hsize=0.55\hsize\raggedright\arraybackslash}X
  >{\hsize=0.87\hsize\raggedright\arraybackslash}X
  >{\hsize=0.83\hsize\raggedright\arraybackslash}X
  >{\hsize=1.5\hsize\raggedright\arraybackslash}X
  >{\hsize=0.25\hsize\centering\arraybackslash}X
@{}}
\toprule
\textbf{System} & \textbf{Base / Executor Model} & \textbf{Task} & \textbf{Reported Result} & \textbf{OS} \\
\midrule

\rowcolor{catFeedback}\multicolumn{5}{@{}l}{\textbf{\faSync~Execution Feedback}}\\
\addlinespace[1pt]
\textbf{SkillForge}~{\footnotesize\cite{liu2026skillforge}}
  & Qwen3-Max
  & Cloud technical support
  & Self-evolution loop gains $+10.99$ / $+9.23$ / $+11.60$\,pp Strict Consistency Rate; $+13.76$\,pp over production legacy system
  & Yes \\
\textbf{CoEvoSkills}\newline{\footnotesize\cite{zhang2026coevoskills}}
  & Claude Opus 4.6
  & SkillsBench (11 domains)
  & 71.1\% pass rate ($+40.5$\,pp over no-skill); transfers to 6 LLMs
  & Yes \\
\textbf{Skills-Coach}~{\footnotesize\cite{tian2026skills}}
  & claude-sonnet-4-6 (training-free GRPO)
  & Skill-X (48 skills)
  & Pass rate $33.59\% \!\to\! 88.02\%$ ($+54.43$\,pp); avg.\ score $0.378 \!\to\! 0.84$ on Skill-X
  & Yes \\
\textbf{Ctx2Skill}~{\footnotesize\cite{si2026context}}
  & GPT-4.1 / 5.1 / 5.2
  & CL-bench (context learning)
  & Consistently lifts solving rate: GPT-4.1 $11.1\% \!\to\! 16.5\%$, GPT-5.1 $21.2\% \!\to\! 25.8\%$
  & Yes \\
\textbf{AutoSkill}~{\footnotesize\cite{yang2026autoskill}}
  & Model-agnostic plug-in layer (training-free)
  & WildChat-1M dialogue traces
  & No task-accuracy eval; descriptive SkillBank analysis: 1{,}858 skills from 22{,}511 conversations
  & Yes \\
\textbf{SkillClaw}~{\footnotesize\cite{ma2026skillclaw}}
  & Qwen3-Max
  & WildClawBench (4/6 categories; 8 users, 6 rounds)
  & Relative gains over Day 1 after 6 rounds: Creative $11.57 \!\to\! 21.80$, Search $22.73 \!\to\! 34.55$, Safety $24.00 \!\to\! 32.00$, Social $54.01 \!\to\! 60.34$
  & Yes \\
\textbf{EmbodiSkill}~{\footnotesize\cite{ju2026embodiskill}}
  & Qwen3.5-27B / Qwen3-VL-32B
  & ALFWorld, EmbodiedBench
  & 93.28\% ALFWorld ($+31.58$\,pp over GPT-5.2 direct agent)
  & No \\
\midrule

\rowcolor{catDistill}\multicolumn{5}{@{}l}{\textbf{\faRoute~Trajectory Distillation}}\\
\addlinespace[1pt]
\textbf{SPARK}~{\footnotesize\cite{zhou2026evidence}}
  & Claude Opus 4.6\newline GPT-5.4
  & SkillsBench (86 tasks) + 300 variants; ALFWorld (OOD)
  & Pass rate $+12.1$ to $+33.3$\,pp (e.g.\ GPT-5.1-Codex $0.0\% \!\to\! 24.2\%$); mean reward beats human skills (0.52 vs 0.47)
  & Yes \\
\textbf{Trace2Skill}~{\footnotesize\cite{ni2026trace2skill}}
  & Qwen3.5-35B, Qwen3.5-122B
  & Spreadsheet, math, VisionQA, WikiTableQuestions
  & Skills from Qwen3.5-35B traces improve a Qwen3.5-122B agent by up to $57.65$\,pp on WikiTableQuestions
  & Yes \\
\textbf{Memento-Skills}~{\footnotesize\cite{zhou2026memento}}
  & Gemini-3.1-Flash, Qwen3-Embedding-0.6B router
  & GAIA (165 q) and HLE (1{,}130 q, 8 subjects)
  & GAIA $52.3 \!\to\! 66.0$ ($+13.7$), HLE $17.9 \!\to\! 38.7$ ($+20.8$); library grows $5 \!\to\! 41$ (GAIA) / 235 (HLE)
  & Yes \\
\textbf{XSkill}~{\footnotesize\cite{jiang2026xskill}}
  & Gemini-2.5-Pro / 3-Flash; GPT-5-mini; o4-mini; text-embedding-3-small
  & VisualToolBench, TIR-Bench, AgentVista, MMSearch-Plus, MMBrowseComp
  & Average@4 gains $+2.58$ to $+6.71$\,pp over tool-only baseline (best: Gemini-3-Flash $33.63 \!\to\! 40.34$)
  & Yes \\
\midrule

\rowcolor{catCompAug}\multicolumn{5}{@{}l}{\textbf{\faCompress~Compression \& Augmentation}}\\
\addlinespace[1pt]
\textbf{SkillNet}~{\footnotesize\cite{liang2026skillnet}}
  & Multiple backbones
  & ALFWorld, WebShop, ScienceWorld
  & ${\sim}+40\%$ avg reward, ${-}30\%$ execution steps
  & Yes \\
\textbf{SkillX}~{\footnotesize\cite{wang2026skillx}}
  & Qwen3-32B / Kimi-K2 / GLM-4.6
  & AppWorld, BFCL-v3, $\tau^2$-Bench
  & ${\sim}10$\,pp improvement when plugged into weaker agents
  & Yes \\
\textbf{SkillReducer}\newline{\footnotesize\cite{gao2026skillreducer}}
  & DeepSeek-V3 (compression); Qwen3.5 (eval)
  & 600 skills + SkillsBench; study of 55{,}315 skills
  & 48\% description / 39\% body token reduction, 26.8\% end-to-end savings; quality $+2.8\%$, 86.0\% pass rate
  & No \\
\textbf{SkillFoundry}\newline{\footnotesize\cite{shen2026skillfoundry}}
  & Coding agent (+ Biomni)
  & MoSciBench + genomics
  & 71.1\% mined skills novel; improves 5/6 MoSciBench + 2 genomics tasks
  & Yes \\
\midrule

\rowcolor{catRL}\multicolumn{5}{@{}l}{\textbf{\faTrophy~Reinforcement Learning}}\\
\addlinespace[1pt]
\textbf{D2Skill}~{\footnotesize\cite{tu2026dynamic}}
  & Qwen2.5-7B-Instruct
  & ALFWorld, WebShop, search QA
  & 90.6 overall success on ALFWorld; 84.4 success on WebShop
  & Yes \\
\textbf{SkillRL}~{\footnotesize\cite{xia2026skillrl}}
  & Qwen2.5-7B \newline (+ o3 teacher)
  & ALFWorld, WebShop, 7 search-QA
  & 89.9\% success on ALFWorld; 72.7\% on WebShop
  & Yes \\
\textbf{SkillOS}~{\footnotesize\cite{ouyang2026skillos}}
  & Qwen3-8B
  & ALFWorld, WebShop; AIME24/25, GPQA-Diamond
  & ALFWorld avg success 61.2; WebShop score 40.6, success 16.5
  & No \\
\textbf{Skill1}~{\footnotesize\cite{shi2026skill1}}
  & Qwen2.5-7B-Instruct + frozen all-MiniLM-L6-v2
  & ALFWorld, WebShop
  & ALFWorld avg success 97.5; WebShop score 89.7, success 82.9
  & Yes \\
\bottomrule
\end{tabularx}
\end{table*}